\documentclass[%
 reprint,
 amsmath,amssymb,
 aps,
]{revtex4-2}

\usepackage{graphicx}% Include figure files
\usepackage{dcolumn}% Align table columns on
\usepackage[dvipsnames]{xcolor}
\definecolor{MyRed}{RGB}{236,30,36}
\definecolor{MyBlue}{RGB}{32,119,181}
\definecolor{MyGreen}{RGB}{46,160,72}
\definecolor{MyYellow}{RGB}{244,126,32}

\usepackage{bm}% bold math
\DeclareUnicodeCharacter{2212}{-}
\usepackage[normalem]{ulem}
\usepackage{hyperref}
\usepackage{tabularx}
\usepackage{array}

\begin{document}

\preprint{APS/123-QED}

\title{Reprogrammable sequencing for physically intelligent under-actuated robots }% Force line breaks with \\

\author{Leon M. Kamp\textsuperscript{1}, Mohamed Zanaty\textsuperscript{1}, Ahmad Zareei\textsuperscript{1},  Benjamin Gorissen\textsuperscript{2}, Robert J. Wood\textsuperscript{1}, Katia Bertoldi\textsuperscript{1}}
 % \altaffiliation[Also at ]{John. A. Paulson School of Engineering and Applied Sciences, Harvard University, Cambridge, MA 02138, USA. \CM{Why have this asterisk here if it's the same school? Probably remove}}%Lines break automatically or can be forced with \\
\affiliation{
{\textsuperscript{1}J.A.Paulson School of Engineering and Applied Sciences, Harvard University, USA.}
}
\affiliation{
{\textsuperscript{2}Department of Mechanical Engineering, KULeuven and Flanders Make, Belgium.}
}

\begin{abstract}
Programming physical intelligence into mechanisms holds great promise for machines that can accomplish tasks such as navigation of unstructured environments while utilizing a minimal amount of computational resources and electronic components. In this study, we introduce a novel design approach for physically intelligent under-actuated mechanisms capable of autonomously adjusting their motion in response to environmental interactions. Specifically, multistability is harnessed to sequence the motion of different degrees of freedom in a programmed order. A key aspect of this approach is that these sequences can be passively reprogrammed through mechanical stimuli that arise from interactions with the environment.  To showcase our approach, we construct a four degree of freedom robot capable of autonomously navigating mazes and moving away from obstacles. Remarkably, this robot operates without relying on traditional computational architectures and utilizes only a single linear actuator.
\end{abstract}

\maketitle
\section*{Introduction}

Autonomous interactions with unstructured environments pose significant challenges for robots, often necessitating complex perception and controls systems, and a multitude of sensors and actuators.  However, there have been investigations in recent years that demonstrated how the incorporation of ``physical intelligence'' directly into the body of a robot can result in autonomous responses to environmental cues while utilizing fewer sensors, actuators, and controllers~\cite{Sitti2021,drotman2021electronics}. It has been shown that under-actuated mechanisms provide a promising framework to realize physically intelligent systems that can use additional degrees of freedom to adapt to changing boundary conditions. For example, under-actuated mechanisms with specific regions of tuned compliance have resulted in cockroach-inspired robots capable of robust locomotion through uneven terrains~\cite{Clark2001} and robotic grippers that can efficiently grasp objects of different sizes, shapes, and stiffness levels using a single actuator~\cite{Dollar2010, Odhner2014, Catalano2014}. Furthermore, programmable compliance in under-actuated mechanism  has been shown to simplify control in limbless robots~\cite{Wang2023}. However, this passive compliance does not allow for the control necessary for coordinated actuation between joints.  For example, although studies have demonstrated that underactuated mechanisms with selective distributed compliance can yield self-stabilizing and energy-efficient walking gaits, these mechanisms still depend on two actuators to synchronize swing and lift degrees of freedom~\cite{badri2022birdbot,Iida2004}. A few examples have been reported of  robots capable of various gaits with just one actuator,  \cite{Zarrouk2014,Hariri2019,Noji2022, Feshbach2023}, but they  still depend on traditional computational frameworks for sensing and control.  

Recently, there have been advances in harnessing multistability to transition between states with distinct programmed behaviors~\cite{Cao2021,Xu2023,Osorio2022,jin2020,shan2015,Hanna2014,melancon2021multistable}. This opened new avenues for locomotion without traditional controllers.
For instance, the snap through instability of a curved strip has been utilized to create self-rolling robots capable of autonomously navigating mazes~\cite{Zhao2022}, origami-inspired multiplexed switches have been implemented to realize an untethered crawler that avoids obstacles~\cite{Yan2023}, and bistable mechanical valves have played a crucial role in the development of pneumatic circuits that control the locomotion of soft-legged robots~\cite{Drotman2021}. Extending beyond locomotion, arrays of multiple bistable units have facilitated the encoding of mechanical logic and metamaterials with reprogrammable properties~\cite{Kuppens2021,Waheed2020,preston2019soft,chen_pauly_reis_2021}. Furthermore, by manipulating the energy landscapes of individual degrees of freedom, predetermined transitions between states have been demonstrated~\cite{Hyatt2023,Melancon2022, van2023nonlinear, Sun2024}. The theoretical framework describing these transitions has also been established  \cite{Hecke_theory21}, laying the groundwork for designing multistable corrugated sheets capable of navigating intricate transition pathways~\cite{hecke_2021}. Crucially, it has also been shown that these pathways can be finely adjusted by modulating the systems boundary conditions~\cite{hecke_2021}. Nonetheless, while this presents an exciting opportunity for creating robotic systems responsive to environmental inputs, concrete functional applications of this capability have yet to be demonstrated.

In this work, we  propose a novel design strategy for physically intelligent under-actuated mechanisms with reprogrammable behaviors and harness this to realize robots capable of adapting their gait in response to mechanical interactions with the environment. More specifically, we exploit geometric nonlinearities combined with elasticity to create reprogrammable mechanisms with multi-welled energy landscapes that yield a variety of minimum energy pathways with a single actuator input. Furthermore, we demonstrate the tunability of the energy landscape using mechanical inputs to autonomously reprogramreprogram the motion sequence of the mechanism.  As a practical demonstration of this framework, we  create a four degree of freedom robot capable of navigating mazes and avoiding obstacles---all without the need for computational intelligence and utilizing only a single actuator.

\section*{Results}

\begin{figure}[t]
\includegraphics[width =1.0\linewidth]{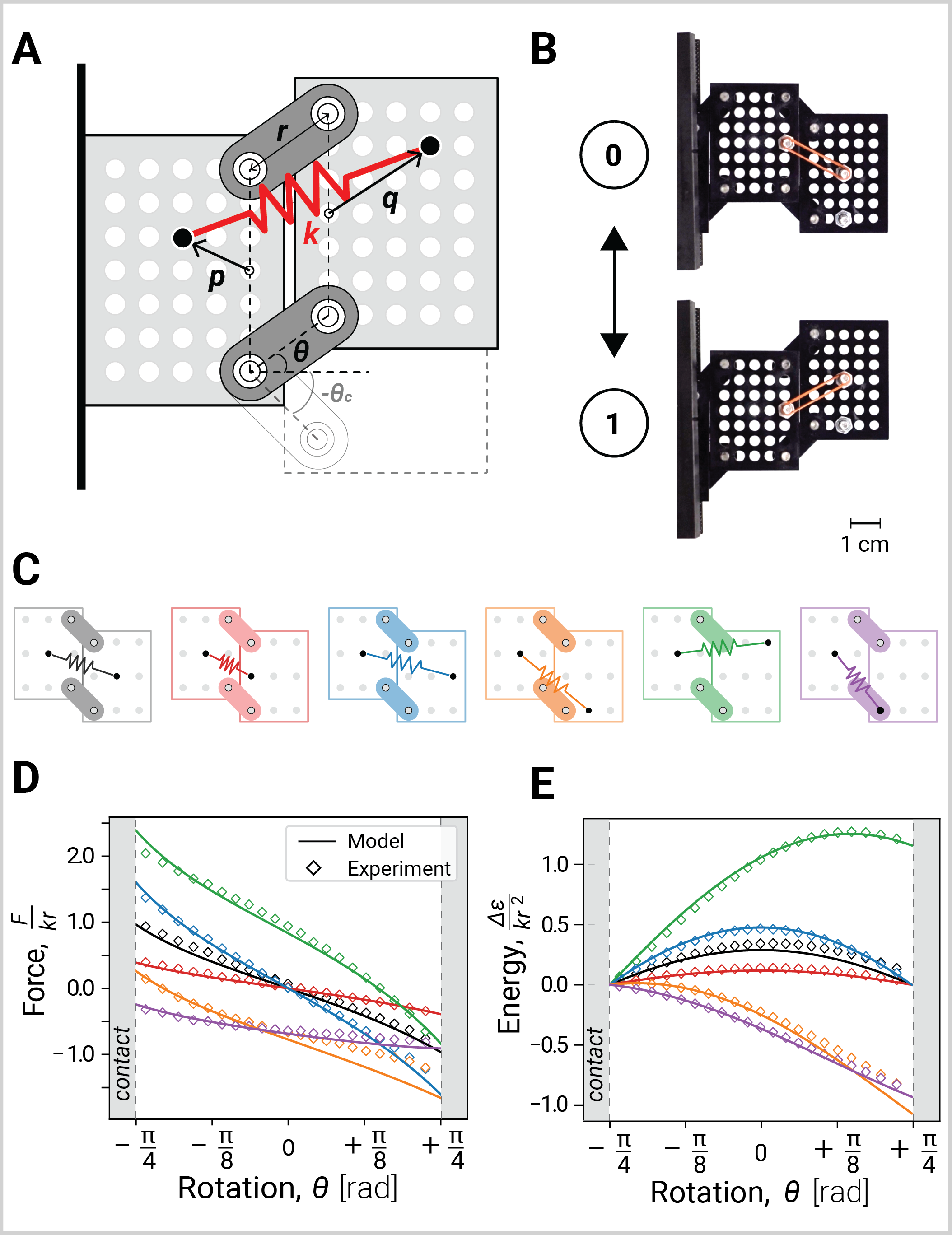}% Here is how to import EPS art
\caption{(A) Schematic representation of a unit cell. (B) Picture of a unit cell in its two possible stable states, \emph{state 0} and \emph{state 1}. (C) Schematic of a  unit cell with six different rubber band configurations, all defined by  $\mathbf{p}$ = [-1, 0] cm, but varying 
$\mathbf{q}$ = [10,0], [0,0], [20,0], [10,-15], [20,15] and [0,-15] mm (black, red,blue, yellow, green, purple, respectively). (D)-(E) Evolution of the  vertical (D) reaction force $F$ and (E) strain energy $\Delta\mathcal{E}$ as a function of the rotation of the levers $\theta$. The colors correspond to the configurations shown in (C).}
\label{fig:single unit_main}
\end{figure}

\paragraph{Characterization of the unit cell. } 
Our unit cell with one degree of freedom, consists of a parallelogram four-bar mechanism comprising two rectangular blocks connected by a pair of identical levers of length $r$ and a spring (rubber band). As shown in Fig.~\ref{fig:single unit_main}A, the two ends of the rubber band are anchored  at points defined by the vectors $\mathbf{p}$ and $\mathbf{q}$, which   originate from the midpoints between the joints of the levers. The range of rotation  of  the levers is constrained by the contact between the blocks on the interval [$-\theta_c$, $\theta_c$]. We define the configurations where $\theta=-\theta_c$ and $\theta_c$ as \emph{state 0} and \emph{state 1}, respectively (with $\theta$ representing the angle between the levers and the horizontal direction). Modeling the rubber band  as a linear spring with stiffness $k$ and rest length $\ell_0$, the energy landscape of the unit cell is given by
\begin{align}
\label{energy}
    \mathcal{E}(\theta,\mathbf{p},\mathbf{q}) = \frac{k}{2}\left[\ell(\theta,\mathbf{p},\mathbf{q}) - \ell_0\right]^2,
\end{align}
with
\begin{equation}
\ell(\theta,\mathbf{p},\mathbf{q})=\sqrt{(r\cos\theta + q_x -p_x)^2 + (r \sin \theta + q_y - p_y)^2},
\end{equation}
where  ($p_x,\,p_y$) and ($q_x,\,q_y$) denote the $x$ and $y$ components of the $\mathbf{p}$ and $\mathbf{q}$ vectors. 

To explore how the location of the anchor points of the rubber band affects the mechanical response, we consider six unit cell designs with varying $\mathbf{q}$, but constant $\theta_c=\pi/4$, $r=\sqrt{2}$ cm, $p_x= -1$ cm, $p_y= 0$ cm, $k = 28.5$ N/m and $\ell_0= 10$ mm (Fig.~\ref{fig:single unit_main}C). All units are placed in 
\emph{state 0} and moved to \emph{state 1} by applying an upward displacement, $u$, with magnitude $u_{max} = 2r \sin \theta_c$ to their outer (right) block. In Fig.~\ref{fig:single unit_main}D and~\ref{fig:single unit_main}E  we report the  evolution of the vertical reaction force on the outer block, $F$, and the difference in energy between the current configuration and \emph{state 0}, $\Delta\mathcal{E} = \mathcal{E}(\theta) - \mathcal{E}(-\theta_c)$, as a function of $\theta$. Three key features emerge from these plots. Firstly, for all units the reaction force monotonically decreases as a function of $\theta$,  leading to a concave down energy landscape. Secondly, the location of the anchoring points for the rubber band  strongly affect the energy costs for switching from \textit{state 0} to \textit{state 1}. This influence opens up possibilities for creating reprogrammable sequences within arrays of coupled units. 
Thirdly,  all of the units, except for $\mathbf{q} = [0,-15]$ mm (purple), are bistable. These cases display two energy minima at $\theta=\pm \theta_c$ separated by an energy 
barrier. Note that, since $\mathcal{E}$ is characterized by a local maximum at

\begin{equation}
\theta^{max}=\arctan\left(\frac{p_y-q_y}{p_x-q_x}\right),
\label{theta_max}
\end{equation}
any units with $|\theta^{max}|< \theta_c$ will display bistability and be characterized by two local energy minima located at \textit{state 0} and \textit{state 1}. Conversely, when $\mathbf{p}$ and $\mathbf{q}$  are selected so that $|\theta^{max}|> \theta_c$, we expect the units to have only one stable configuration (either at $\theta=-\theta_c$ or at $\theta=\theta_c$).

To validate these  predictions, we built a prototype comprising laser cut acrylic blocks and levers connected by pin joints with ball bearings. In our experiments we clamp the left block and pull the right from \emph{state 0} to \emph{state 1}, while recording the reaction force using a Instron 5969 equipped with a 500 N load cell. In Fig.~\ref{fig:single unit_main}D and E we compare the experimentally-measured forces and elastic energy (obtained by numerically integrating the measured reaction force) to numerical predictions for the six unit cell designs, with agreement sufficient to validate our simple analytical model.\\

\begin{figure*}[ht]
\includegraphics[width =1.0\linewidth]{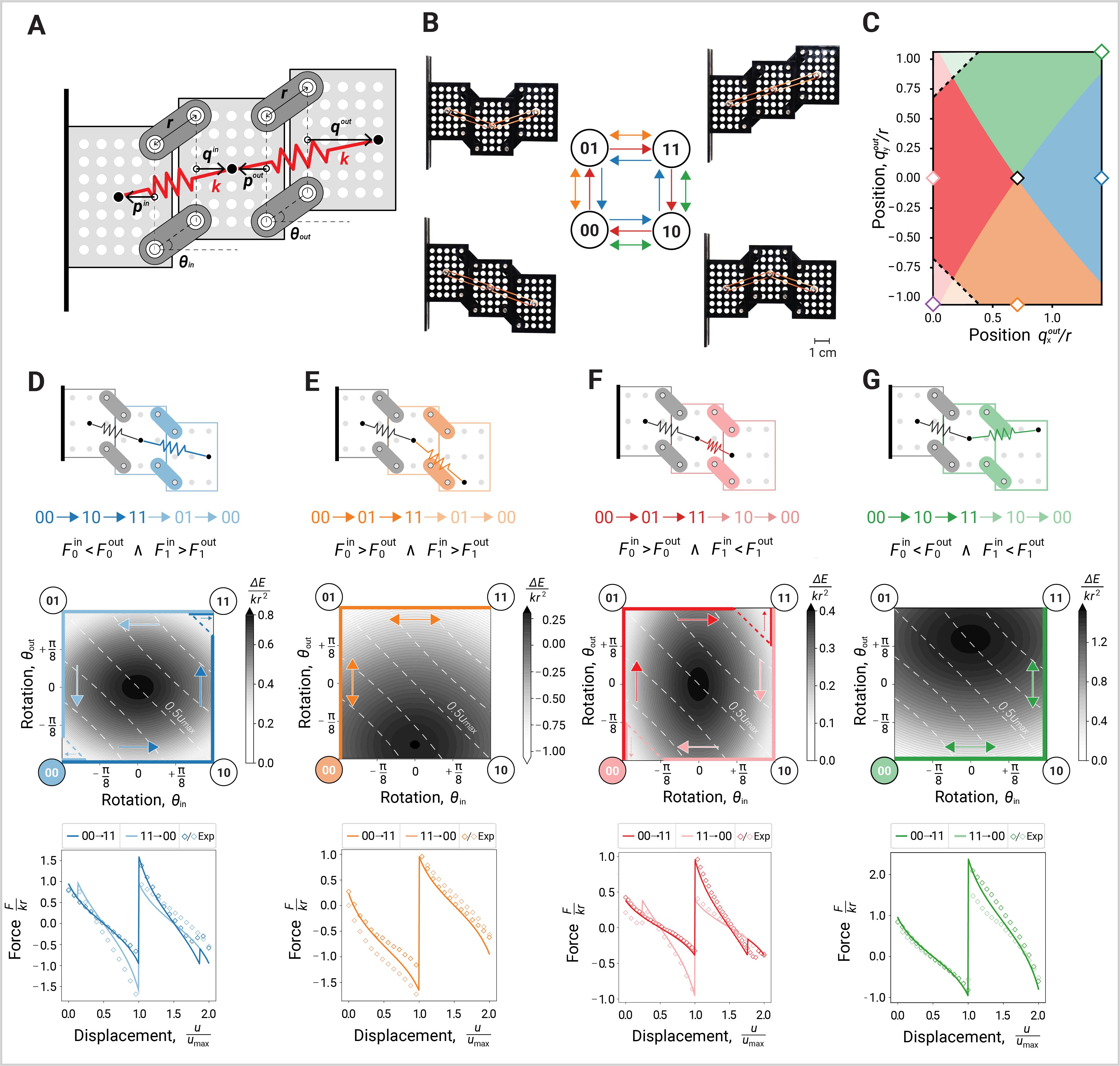}% Here is how to import EPS art
\caption{\label{fig:epsart}
(A) Schematic of a mechanism comprised of two unit cells connected in series. (B) The four possible states and diagram highlighting the four observed transition sequences. (C) Effect of $\mathbf{q}^{out}$ on the sequence of transitions for the mechanism in (A) with $\mathbf{p}^{in} = −\mathbf{q}^{in} =$ [−10, 0] mm. The colors correspond to the sequences highlighted in  (B) and  the light shaded areas denote  unit cells that are  monostable when  considered individually. The markers highlight mechanisms built choosing from the five different configurations of the unit cell shown in Fig. \ref{fig:single unit_main}C as the outer unit.  (D)-(G) We consider  four of the two-unit mechanisms indicated by a marker in (C) and for each of them show a schematic (top), the energy landscape together with the  supported state transitions (center), and the force-displacement response measured in experiments and predicted by theory.} 
\label{fig:two_units_main}
\end{figure*}

\paragraph{Serial coupling of two unit cells. }Next, with the goal to encode deformation sequences, we turn our attention to a mechanism comprised of two unit cells connected in series: an inner unit on the left and an outer unit on the right (Fig.~\ref{fig:two_units_main}A). The state of this system is characterized by the unit states ($\alpha^{in} \alpha^{out}) \in \{0,1\} $ of the inner and outer unit, respectively. Starting at state  (00), we apply an upward  vertical displacement of magnitude $2 u_{max}$ to the outermost block that causes a transition to state (11) and then return to state (00) by applying a  downward displacement of identical magnitude. The sequence of transitions connecting states (00) and (11) can be deliberately controlled by manipulating the  energy landscape of the two units. By carefully selecting the anchoring points of the rubber bands, defined by the vectors ($\mathbf{p}^{in}$, $\mathbf{q}^{in}$) and ($\mathbf{p}^{out}$, $\mathbf{q}^{out}$), we have the capability to engineer directed pathways that cyclically traverse all four possible states under a linear input (blue and red arrows in Fig. \ref{fig:two_units_main}B). Additionally, we can create undirected pathways that access the states in the same order when applying upward and downward displacement (green and orange arrows in Fig. \ref{fig:two_units_main}B). Note that for the considered mechanism, it is impossible to switch two states simultaneously (e.g., transitioning from state (01) to state (10)).
 
To determine the pathway followed by the system, we first define its
elastic energy,   which is given by the sum of the elastic energy of the individual units
\begin{equation}
    \mathcal{E}_{tot}=\mathcal{E}(\theta^{in},\,\mathbf{p}^{in},\, \mathbf{q}^{in})+\mathcal{E}(\theta^{out},\,\mathbf{p}^{out},\, \mathbf{q}^{out}),
\end{equation}
where $\theta^{in}$ and $\theta^{out}$    represent the angles between the horizontal direction and the levers of the inner and outer unit, respectively. When the displacement of the outer unit is controlled, the response of the system is characterized by one independent variable and the constraint 
\begin{equation}
u=r(\sin\theta^{in}+\sin\theta^{out}-2 \sin\theta_c).
\end{equation}
To identify the path followed by  the structure,   we incrementally increase $u$ starting from the initial configuration (defined by $\theta^{in}=\theta^{out}=-\theta_c$) and locally minimize the elastic energy, $\mathcal{E}_{tot}$, using a quasi-Newton method (see Supporting Information for details).

In Fig.~\ref{fig:two_units_main}C  we focus on a mechanism comprised of two unit cells identical to those considered in Fig.~\ref{fig:single unit_main}. We choose $\mathbf{p}^{in}=-\mathbf{q}^{in}=\mathbf{p}^{out}=[-10,\,0]$ mm, and systematically investigate the effect of $\mathbf{q}^{out}$ on the sequence of transitions.   We find  that large values of ${q}_y^{out}$ tend to promote undirected transition sequences (orange and green shaded areas in Fig.~\ref{fig:two_units_main}C),  whereas large values of ${q}_x^{out}$ tend to lead to directed transition sequences (red and blue shaded areas in Fig.~\ref{fig:two_units_main}C). In  Figs. \ref{fig:two_units_main}D-\ref{fig:two_units_main}G we examine four configurations, with $\mathbf{q}^{out}$=[0,0], [20,0], [10,-15], [20,15] mm, that display the four supported sequences of transitions (corresponding to the diamond markers in Fig.~\ref{fig:two_units_main}C). For each of them we report the energy landscape as a function of $\theta^{in}$ and $\theta^{out}$, alongside the measured vertical reaction force at the outer unit as a function of the applied displacement.  For all configurations, we observe that the minimum energy path results in sequential motions, where the units  transition between \emph{state 0} and \emph{state 1} one after the other, rather than simultaneously. This is  due to the convex down nature of the energy landscape of the individual units, resulting in energy minima  predominantly  along the  boundary. For this reason, the order of these sequential transitions can be determined by the reaction force of the two individual units at \emph{state 0}, $F_0=F(-\theta_c)$, and at \emph{state 1}, $F_1=F(\theta_c)$. More specifically, undirected pathways that traverse through state (01) are realized when $F_0^{in}>F_0^{out}$ and $F_1^{in}>F_1^{out}$, while those transitioning through state (10) require $F_0^{in}<F_0^{out}$ and $F_1^{in}<F_1^{out}$.  Conversely, clock-wise directed sequences  necessitate that $F_0^{in}<F_0^{out}$ and $F_1^{in}>F_1^{out}$, while  the counterclockwise sequences require $F_0^{in}>F_0^{out}$ and $F_1^{in}<F_1^{out}$. We also note that for all the configurations considered here supporting undirected sequences, the system remains exclusively along the boundaries of the energy landscape.  In contrast, a snapping instability is triggered for  all configurations supporting directed sequences, when approaching states (11) and (00), (see Supporting Information for details). 
 Finally, it is noteworthy that the combined energy landscape of all the two-unit mechanisms examined here reveals a multi-welled landscape, even in situations where the outer units are monostable (represented by lightly shaded regions in Figure~\ref{fig:two_units_main}C).\\

\begin{figure*}[ht]
\includegraphics[width =1.0\linewidth]{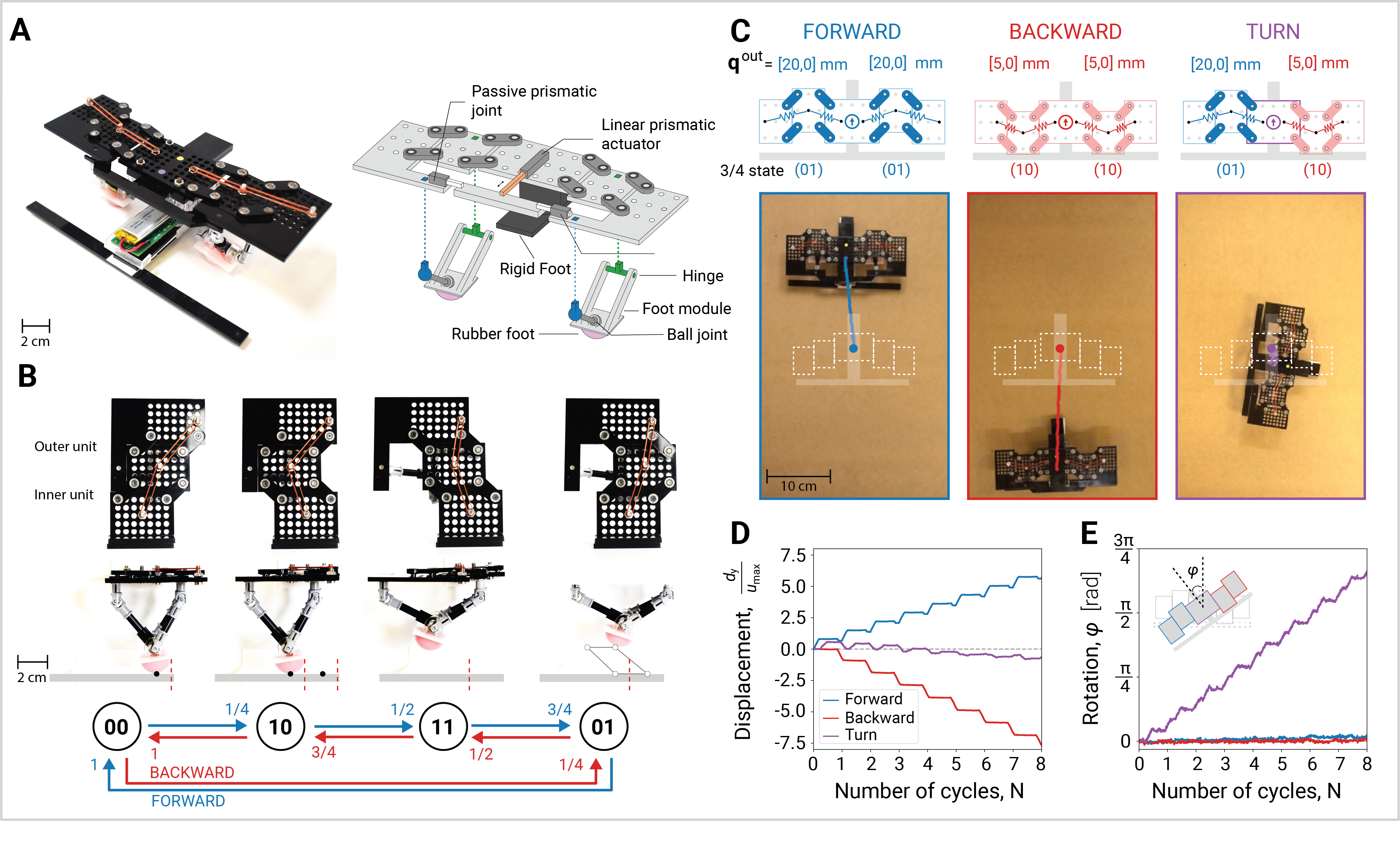}
\caption{(A) Picture and schematic of the under-actuated robot. (B) Top and side view of one leg of the robot as it moves through its four supported states. Moving the inner unit from
\emph{state 0} to \emph{state 1} propels the entire mechanism forward, whereas shifting the outer unit from \emph{state 0} to \emph{state 1} raises the foot. (C) Trajectory followed by the robot for $\mathbf{q}^{out} = [20,0]$ mm (left) and   $\mathbf{q}^{out} = [5,0]$ mm (center) on both sides, and  $\mathbf{q}^{out} = $[20,0] mm and  $\mathbf{q}^{out} = $[5,0] mm on the left and right side, respectively (right). The initial position corresponds to the white dashed outlines, while the picture indicate the position of the robot 
after $7\frac{3}{4}$ cycles. (D)-(E) Evolution of the (D) displacement, $d$, and rotation, $\varphi$, of the
robot for the three  configurations of the rubber bands considered in (C).}
\label{fig:robot_main}
\end{figure*}

\begin{figure*}[ht]
\includegraphics[width =\linewidth]{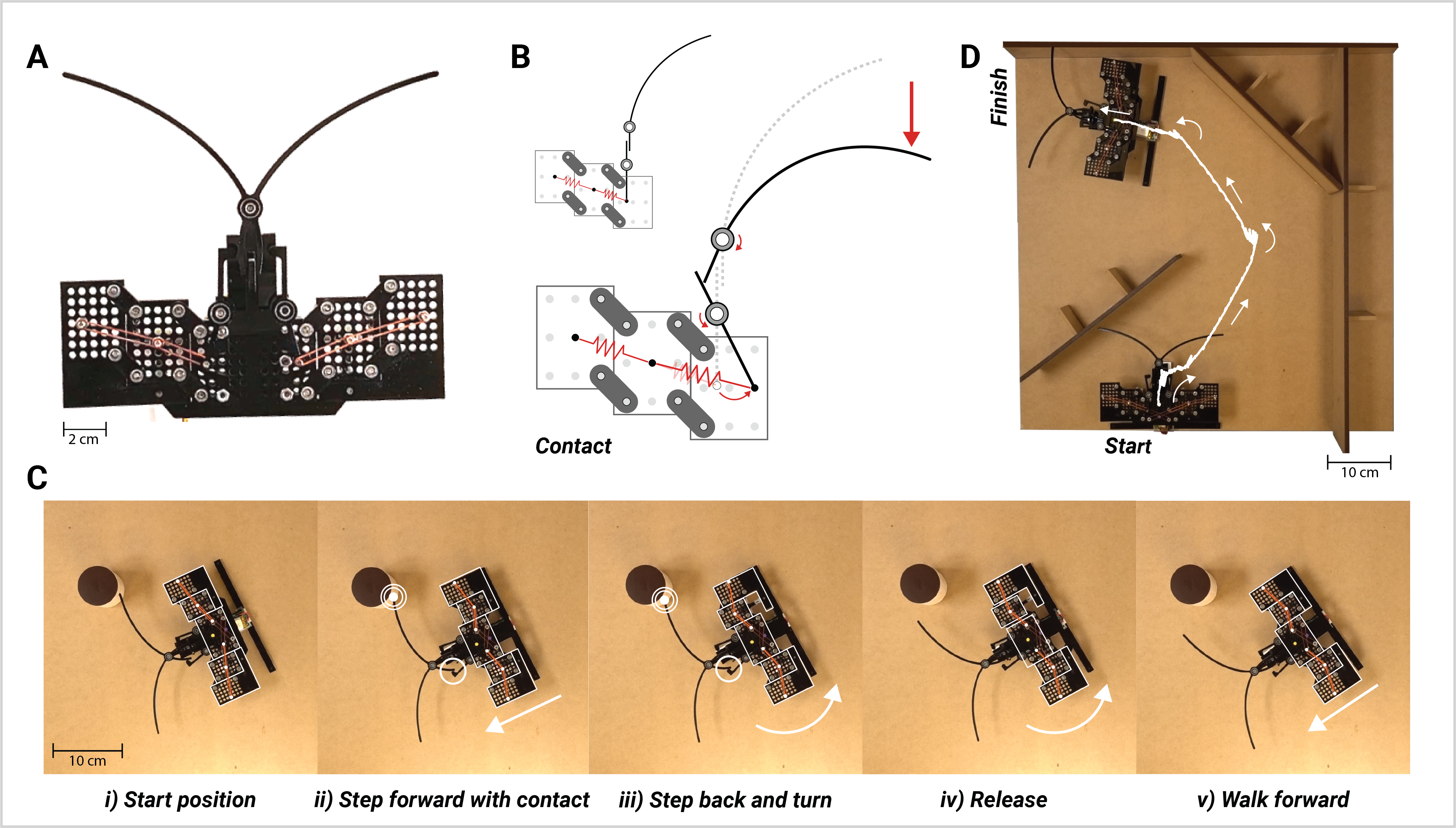}
\caption{(A) Picture of the robot with  a module with two antennas
attached at its front. (B) Schematic of showing that the movement of the right antenna
alters $\mathbf{p}^{in}$ of the left mechanism. (C) Snapshots of the robot interacting with a cylindrical obstacle (see also Video S4) (D) Trajectory of
the robot navigating through an environment obstructed
by three walls.}
\label{fig:Robot_sensing}
\end{figure*}

\paragraph{Physically intelligent under-actuated robot.}
Next, we demonstrate how to encode various transition sequences in an under-actuated three-legged robot to realize multiple modes of locomotion. This robot controls four degrees of freedom with a single actuator and can be physically reprogrammed to change gait. As shown in Fig.~\ref{fig:robot_main}A, the robot is comprised of a pair of the two-unit mechanisms described in Fig.~\ref{fig:two_units_main}, two  V-shaped leg mechanisms, and a body with a stepper motor, battery, controller and a large T-shaped foot for stability. The innermost block of the two-unit mechanism is anchored to the rigid body, while the stepper motor drives the outermost block back and forth, simultaneously driving the two mechanisms back and forth between their (00) and (11) states. We connect the V-shaped leg mechanism to both the middle and outermost blocks of the two-unit mechanisms. As a result, shifting the outer unit from state 0 to state 1 increases the distance between the connection points of the leg, raising the foot. Conversely, a transition from state 1 to state 0 lowers the foot. Moving the inner unit from state 0 to state 1 propels the entire mechanism forward, while transitioning from state 1 to state 0 pushes the mechanism backward (Fig. \ref{fig:robot_main}C). Therefore, the sequence of transitions (00) $\rightarrow$  (10) $\rightarrow$ (11) $\rightarrow$  (01) $\rightarrow$(00) leads to a forward step, whereas (00) $\rightarrow$  (01) $\rightarrow$ (11) $\rightarrow$  (10) $\rightarrow$(00) results in a backward step (Fig. \ref{fig:robot_main}C).  

 We carefully  select  the anchor points of the rubber bands to ensure the motion of the leg follows the desired sequence of transitions. It is important to recognize that the robot's response is also influenced by external factors, unlike a standalone mechanism where the energy landscape is solely determined by the springs (i.e., their stiffness, initial length, and anchoring points). Specifically, we need to account for the effects of friction on the inner unit and gravity on the outer unit (see Supporting Information for details). In our robot, we observe that the contribution of these forces is significant enough to affect the transition map of Fig.\ref{fig:two_units_main}C. For instance, while a standalone mechanism with anchor points defined by $\mathbf{p}^{in}=$[-10, 0] mm, $\mathbf{q}^{in}=$[10, 0] mm, and $\mathbf{p}^{out}=$[-10, 0] mm achieves a sequence leading to a forward step for $\mathbf{q}^{out}=$[15, 0] mm (as indicated by the blue region in Fig.\ref{fig:two_units_main}C), for the robot, this is only feasible when $\mathbf{q}^{out}$ is adjusted to [20, 0] mm (see Video S3 and Supporting Information for details).  Conversely, the sequence leading to a backward step is less affected by  friction as contact with the ground is only made in the second part of the sequence (i.e., $(10) \rightarrow (00)$). Therefore, a backward step can be realized by selecting $\mathbf{q}_{out}$=[5, 0] mm (see Video
S3 and Fig. S13B).  
In Fig.~\ref{fig:robot_main}D, we show the trajectory followed by the robot for different choices of $\mathbf{q}^{out}$. As expected, we see that the robot moves forward for $\mathbf{q}^{out}=$[20, 0] mm (left), and moves backward for $\mathbf{q}^{out}=$[5, 0] mm (center). Further, a turning motion is achieved by programming one leg to execute a forward step while the other performs a backward step.
The difference between a forward, backward, and turning motion is also captured in the state of the mechanism. As shown in Fig. \ref{fig:robot_main}C, the leg with $\mathbf{q}^{out}=$[20, 0] mm is in state (01) after completing $7\frac{3}{4}$ input cycles, whereas the one with $\mathbf{q}^{out}=$[5, 0] mm  is in state (10).
Moreover, in Figs.~\ref{fig:robot_main}E-F, we present the evolution of the displacement, $d$, and rotation, $\phi$, of the robot for the three considered configurations of the rubber bands. We observe that for the robot programmed to move backward, $d/u_{max} = 0.96$, indicating that the displacement provided by the linear actuator is almost entirely converted into movement of the robot. However, for the forward motion, $d/u_{max} = 0.71$. We attribute this reduction to the snapping of the mechanism as it approaches the (11) state, exerting a force that pushes the robot in the opposite direction. Lastly, we find that, for turning motion,  $d$ is negligible and $\phi \approx \pi/12$ rad per cycle. \\

\paragraph{Gait adaptation in response
to mechanical interactions with the environment.}
In Fig.~\ref{fig:robot_main} we show that a small linear displacement of an anchor point can completely reverse the gait of a leg. We harness this observation to realize a robot capable of adapting its gait in response
to mechanical interactions with the environment.  For this, we attach a module with two antennas at the front of the robot (Fig. \ref{fig:Robot_sensing}A). Each antenna consists of two levers that connect to the inner rubber band of the two-unit mechanism on the opposite side of the robot (Fig. \ref{fig:Robot_sensing}B). This makes it possible to change $\mathbf{p}^{in}$ when an antenna comes into contact with an obstacle. More specifically, the movement of  the left (right) lever  alters $\mathbf{p}^{in}$ of the right (left) mechanism. To avoid an obstacle, we  tune the rubber bands so that this change in $\mathbf{p}^{in}$ is enough to make the robot to  turn away from the obstacle.  To allow the levers to move with a low contact force and therefore increase their sensitivity, we use rubber bands with rest length $\ell_0 = 22$ mm and stiffness $k$= 47 N/m. Further, we choose $\mathbf{q}^{in} = [10,0]$ cm, $\mathbf{p}^{out} = [-10,0]$ cm, and $\mathbf{q}^{out} = [20,0]$ mm and find  that that the robot moves forward in the absence of contact if the antennas are tuned to hold the anchoring point of the inner rubber bands at a position defined by $\mathbf{p}^{in} = [-7,0]$ mm (Fig.~\ref{fig:Robot_sensing}C). When the right antenna comes in contact with an obstacle (here, a wooden cylinder with a diameter of 50 mm), it pushes $\mathbf{p}^{in}$ of the left mechanism to  $\approx[-11,0]$ mm . This  initiates a backwards left turning motion until the antennas lose contact with the object. Subsequently, $\mathbf{p}^{in}$ springs back to $[-7,0]$ mm, and the robot resumes its forward motion. The robot's ability to move away from mechanically sensed obstacles is observable for a flat wall and cylindrical column across a wide range of approaching angles and distances from the center of the robot (see Video S4).  As a demonstration of the robustness of the observed behavior, in Fig.~\ref{fig:Robot_sensing}D we depict the trajectory of the robot navigating through an environment obstructed by three walls. The robot is able to successfully traverse this environment and reach the exit without electronic sensing and using a single actuator (Video S4).

\paragraph{Conclusions.}
To summarize, we have shown that a multistable energy landscape enables us to create forward and backward locomotion gaits with a single quasi-static linear input. The tunable nature of the energy landscape at each degree of freedom makes it possible  to realize a robot capable of adapting its gait
in response to mechanical interactions with the environment without the need for electronic feedback and control.   Though our focus in this study has been on mechanisms with two degrees of freedom, our approach can readily extend to systems with a greater number of degrees of freedom, thereby expanding the range of possible states and transitions. This consequently allows for more complex control strategies and mechanical computing~\cite{Kuppens2021,Waheed2020,preston2019soft,Hyatt2023, Yasuda2021}. Furthermore, while our study has centered on a platform consisting of rigid blocks connected by levers and elastic springs, multistable energy landscapes can also be realized using beams and shells, which allow for monolithic fabrication~\cite{Gou2021,TENWOLDE2024105626,Gorissen2020,Osorio2022}. This opens up avenues for potential integration into robotic systems with size, weight or material constraints for on-board control, such as soft robots and microrobots~\cite{Dolev2021,Pal2021,Bandari2021}. Finally, given the potential for triggering snapping instabilities in multistable mechanisms, there is an opportunity for leveraging this phenomenon for rapid movements such as jumping over obstacles~\cite{Gorissen2020, Carlson2020}.
All together, our platform provides a foundation for designing physically intelligent autonomous robots that operate with minimal reliance on electronic controllers, sensors, and actuators.

\section*{Materials and methods}
Details of the design, materials, fabrication, testing methods and mathematical model of the unit cell and the two-unit mechanism are summarized in Section S1 of Supporting Information. The design, fabrication and analysis of the robot is discussed in Section S2 of Supporting Information.

\section*{Acknowledgements}
The authors gratefully acknowledge support from the National Science Foundation through the Harvard MRSEC (DMR-2011754) and the ARO MURI program (W911NF-22-1-0219). The authors thank Connor McCann, Michelle Yuen, Colter Decker, Adel Djellouli, Anne Meeussen, Davood Farhadi and Giovanni Bordiga for helpful discussions.

\section*{Data availability statement}
The source code developed for the analytical model and post-processing of experimental data is available on GitHub at \textit{https://github.com/kampleon/Sequence\_robot/}. All the processed is attached as Supporting Information. Raw data is available on the GitHub repository.

\section*{Competing interests} The 
authors declare that they have no competing interests.

%While in thsi study we have focused on mechanism with two degrees of freedom, the proposed apprioach can be reasily extended to sytems with largeer number of degree of freedom to further increase  the number of possible states and transitions, and therfore potential functionelity. Further since multistable mechanism snapping instabilities can be triggered, thsi provides an opportunity to harness   harness energy storage and release. In this paper the effect of snapping is limited, but could in future works be harnessed to access rapid motions for jumping over or breaking obstacles ~\cite{Pal2020,Steinhardt2021, Sun2024}. Finally, while in theis study we have focused on a  platform comprising rigid blocks connected by lever and elastic springs, multistable energy landscaped can also be realized using beaam and shells. This provide opportunity to monolithis fabrication of them and potential embedding them into rsoft robots and microrobots \LK{need some refs (maybe reviews) for this}. As such, our platform could enable the design of physically intelligent autonomous devices operating with a limited use of electronic controllers, sensors and actuators.


%\bibliography{References}

\end{document}